Review Article

# A review of Implementation and Challenges of Unmanned Aerial Vehicles for Spraying Applications and Crop Monitoring in Indonesia

Authors:

Muhamad Rausyan Fikri[1,2], Taufiq Candra[3], Kushendarsyah Saptaji[4], Ajeng Nindi Noviarini[5], Dilla Ayu Wardani[3]

1. Automation Technology and Mechanical Engineering, Faculty of Engineering and Science, Tampere University, Tampere, 33720, Finland
2. Information Systems, Faculty of Engineering and Technology, Sampoerna University, Jakarta, 12780, Indonesia
3. Industrial Engineering, Faculty of Engineering and Technology, Sampoerna University, Jakarta, 12780, Indonesia
4. Mechanical Engineering, Faculty of Engineering and Technology, Sampoerna University, Jakarta, 12780, Indonesia
5. Computer Science, Faculty of Engineering and Technology, Sampoerna University, Jakarta, 12780, Indonesia

**Abstract:**

The rapid development of technology has brought unmanned aerial vehicles (UAVs) to become widely known in the current era. The market of UAVs is also predicted to continue growing with related technologies in the future. UAVs have been used in various sectors, including livestock, forestry, and agriculture. In agricultural applications, UAVs are highly capable of increasing the productivity of the farm and reducing farmers' workload. This paper discusses the application of UAVs in agriculture, particularly in spraying and crop monitoring. This study examines the urgency of UAV implementation in the agriculture sector. A short history of UAVs is provided in this paper to portray the development of UAVs from time to time. The classification of UAVs is also discussed to differentiate various types of UAVs. The application of UAVs in spraying and crop monitoring is based on the previous studies that have been done by many scientific groups and researchers who are working closely to propose solutions for agriculture-related issues. Furthermore, the limitations of UAV applications are also identified. The challenges in implementing agricultural UAVs in Indonesia are also presented.

**Keywords**:

Unmanned aerial vehicle, agricultural UAV, spraying, crop monitoring.

## 1. Introduction

According to the United Nations (UN), the world population is projected to reach 9.7 billion people in 2050 (UN, 2015). This vast population would potentially double the food demand in the future (Hunter et al., 2017). Consequently, the ever-growing population that would emerge could cause food shortages in the future. This issue has become a severe problem since the Food and Agriculture Organization (FAO) announced similar speculation in which the current agricultural production must be increased by 70 percent by 2050 to meet the increasing demand for high-quality food (Mundial, 2021). Many people suffering from hunger become a signal of how severe the food shortage is, and it was reported that more than 820 million people in 2018 were considered undernutrition (WHO, 2019). Surprisingly, the earlier data mentioned shows the increasing tendency towards people suffering from hunger since only around 690 million people were considered suffering from hunger in 2015. This kind of data indeed contradicts the second Sustainable Development Goals (SDGs) approved by the United Nations (UN) in 2015 with the aims to eradicate hunger and ensure access to food for all people (UN, 2015). On the other hand, the labor shortages in the agricultural sector due to the aging population and the decreasing number of workers have exacerbated the situation. The lack of laborers in the agricultural field would expand the cultivation area per worker and increase the workload of workers (Seo & Umeda, 2021).

Alternative and innovative solutions to increase food production in dealing with those issues are needed. One way to increase food production is to promote the internet of things (IoT), robotics, and artificial intelligence (AI). By shifting the workforce to the technology's utilization, it is expected to solve the labor shortages and improve farmers' skills. This transformation is inseparable from revolutionary industries that constantly bring industrial innovations. Some industrial innovations found in recent years, such as sensor technologies, big data, and artificial intelligence (AI), have been considered as the beginning of the "Industry 5.0" era by the European Commission (EC) (EC, 2021). The emergence of technologies characterized by advanced digitalization is believed to play a significant role in increasing production flexibility and making the value chain more robust so that technology could minimize the farmers' workload and improve the speed and accuracy of the work.

Among the technologies mentioned earlier, unmanned aerial vehicles (UAVs) are one viable way to increase food production. UAVs are less expensive and have contributed to many areas in agriculture, including spraying, weed recognition, and crop monitoring (Mogili & Deepak, 2018). UAVs' timely and reliable information about the production, yield and crop management would become beneficial to ensure food safety and security for stakeholders such as farmers and sales units (Martos et al., 2021). UAV technologies in agriculture could also enable the complete monitoring of crop conditions from the beginning of the growing season until the end of harvest (Silver et al., 2017).

Some leading technologies are possible by implementing UAVs in the agriculture sector. Therefore, this paper focuses on reviewing UAVs applications for spraying and crop monitoring in the agricultural field. Some research results on the use of UAVs in spraying and crop monitoring are discussed thoroughly to highlight the use of UAVs and the characteristics of the farming sector. Some limitations exist during UAVs implementation are also reviewed to reveal the gap of UAV implementation in the agriculture field. The rest of this paper is organized as follows. Section 2 describes UAVs' history and the classification of UAVs. Section 3 describes the application of UAVs focusing on spraying and crop monitoring. Section 4 provides some limitations in adopting UAV technologies in the agriculture sector. In section 5, the challenge in implementing UAVs in Indonesia is discussed. The last section provides the conclusion.

**2. Agriculture in Indonesia and Its Challenge**

**2.1. Current Condition**

Agriculture programs in Indonesia have been a big agenda at the national level, such as National Agenda 21, National Development Programs, and Agricultural and Forestry Revitalization Strategies, encouraging Indonesia to adopt sustainable agriculture. The Central Planning Authority (BAPPENAS), the Ministry of Agriculture, and the Environment Ministry have implemented these ideas. Most of these plans include components suitable for effective environmental management of Indonesian agricultural exports.

The motivations for using these tactics have shifted over time, and they seem to be responding to a variety of distinct trends. First and foremost, Indonesian national plans have prioritized socio-economic objectives above ecologically sustainable ones. Nonetheless, environmental concerns have become more critical, as evidenced by recent reforms and the increasing frequency of ecological issues in strategic documents. Second, strategy papers also show a change in direction as the combination of means changes, with less focus on laws and regulations and more attention to the means for market creation and voluntary methods over time. The tensions between diverse skills and conservation goals and local revenue-generating needs have led to different patterns of success in different states across the country. Significant advancements have been achieved in modernizing agro-environmental rules, made possible by increased information and worldwide best practices. The extent to which environmental hazards pose local or global dangers, the degree of environmental degradation of a particular product, and the availability of legal, enforcement, budgetary, and regulatory capacities for sub-national governments all influence the choice of the policy tool.

For practical reasons, Indonesian policymakers have used a range of mechanisms to minimize agriculture's environmental footprint, including direct regulation, market creation or market modification incentives, voluntary and beneficial solutions, and market modification incentives. Policies are implemented via legislative and regulatory mechanisms, which are probably targeted at plantation states and large farms. It is essential to note the existence of obligatory ISPO standards (in the section on local regulatory instruments), since they have just recently been adopted as a result of voluntary standards being adopted as mandatory. Additional factors that impact policymakers' choices to implement a particular instrument include the potential efficacy of the instrument in comparison to its costs and the capacity of the policymaker to enforce the instrument in the face of likely political opposition. In this respect, implementing regulatory and legislative tools seems to be the most effective method of monitoring prominent investments, such as planting restrictions and the demand for environmental impact assessments. According to the findings of the Indonesian research, foreign pressure had a role in the spread of planting restrictions throughout the country. In addition, the implementation of regulatory instruments may be most effective when their administrative and monitoring costs are already integrated into a current administration, such as indirect product charges for import limitations, which are already embedded into an existing administration.

## 2.2. The Challenge in Indonesia's Agriculture

One of the factors is the limited availability of agricultural land in Indonesia due to land reform, which is widespread in big provinces. As the population rose, so did the need for housing.

As a result, developers exploit a large portion of agricultural land to construct real estate. The growth in the number of people also increased demand for trade and tourism, contributing to increased demand for land. Farmers could not be faulted for selling their farms in this scenario. Farmers were driven to sell their lands due to a lack of knowledge and technology, high agricultural costs, and rising necessities. Farmers in Indonesia with low levels of education have little choice except to work outside the agricultural industry; therefore, those who do not own land are tenant farmers. Food price increases should be a dream come true for farmers, as their revenue would almost certainly rise. Unfortunately, because most farmers in Indonesia are tenant farmers, it has become a boomerang for their wellbeing. The rise in food prices has little effect on the well-being of Indonesian farmers. Their income remains minimal, and they must continue to purchase their basic necessities at market prices. Those who own land have benefited from the growing price. Furthermore, the general public's perception of farmers is that they do not do a good job. The younger generation is interested in non-agricultural jobs, such as parenting a farmer's child. Farmers' regeneration is hampered as a result, and many opt to sell their land to be established as capital or to work in the non-agricultural sector.

As the world's population rises at an alarming rate, agriculture must expand to supply the growing demand for food against all odds. The agriculture sector is the most vulnerable to the impact of integrating fresh, modern innovations in eradicating environmental-related challenges and enhancing the current productivity rate. Now, the question is, how could we possibly do this? Marking a third wave of the "Green Revolution", the concept of precision agriculture with technological help such as Unmanned Aerial Vehicle (UAV) has become popular nowadays in the vast area of agriculture due to its tremendous benefits. Farmers and managers can boost operational efficiency, cut expenses, minimize waste, and improve the quality of crops with the aid of accurate data. Overall, technology has been a key component behind agricultural development and other discoveries brought into the industry.

## 3. Unmanned Aerial Vehicle (UAV)
### 3.1 UAVs History

An unmanned Aerial Vehicle (UAV) is an aircraft with no pilot on board; in other words, it refers to auto-piloted aircraft (Ahmad et al., 2021). The unmanned type of aircraft can be operated in two ways, either by a human operator or autonomously operated under the control of an onboard computer (Pablo et al., 2020). According to the US Department of Defense (DOD), UAV can be described as either a single air vehicle (with equipped surveillance sensors) or a UAV system (UAS) that consists of three to six air vehicles, a ground control station, and support equipment (Gertler, 2012). Furthermore, UAVs are often associated with remote sensing in carrying out their task. This remote sensing is commonly known as UAV remote sensing, which combines UAV and remote sensing technology that can quickly capture information about land, environment, and resources for further data processing (Shi & Liu, 2011). In the US and other developed countries, UAV remote sensing has been applied in many fields such as forestry, environmental protection, land, and military (Xiang & Tian, 2011). UAV remote sensing are used because it could be deployed quickly in repeated times. In addition, they are less costly, safer than piloted aircraft, flexible in terms of flying height, and able to obtain very high-resolution imagery (Yang et al., 2011).

The term unmanned aerial vehicles are also known as remotely piloted aircraft (RPA). Even though the terms UAV and RPA are interchangeable, the term UAV is commonly used by aviation organizations (Santos et al., 2019), while the term RPA is widely used in Europe (Gallardo- Saavedra et al., 2018). Back then, in 1930, UAVs were also known as "Queen Bees" (Vroegindeweij et al., 2014) and were initially used for military purposes (Muchiri & Kimathi, 2016). In 1986, UAVs that work specifically in agricultural contexts were introduced by launching UAVs for Montana's forest fires monitoring and followed by the capture of enhanced image resolution using UAVs in 1994 (Muchiri & Kimathi, 2016). Then, a more complex UAV model was finally developed by Yamaha through "Yamaha RMAX," with the primary function for pest control and crop monitoring application (Mogili & Deepak, 2018). This UAV model is used for pesticide spraying in rice fields of Asia. As opposed to ground-based sprayers, the pesticides deposition of this UAV model is quite similar, but this UAV model is used explicitly for a high-value crop environment (Giles & Billing, 2015).

## 3.2 Classification of UAVs

Generally, there are three types of UAV platforms: fixed-wing, rotary-wing UAVs, and non-wing UAVs (Figure 1). A fixed-wing UAV resembles an airplane and requires a runway or Modelsurface (meadow or road) for take-off and landing (Pederi & Cheporniuk, 2015). This kind of UAV uses thrust and aerodynamic lifting forces to fly. It has a larger size than a rotary-wing model and is mainly used for aerial mapping, spraying, and photography over a wide range of time (Li & Yang, 2012). This UAV type typically lacks hovering while offering high-speed flights for longer durations (Ahmad et al., 2021). The gliding capabilities possessed by fixed-wing aircraft could enable greater flight endurance, allowing them to operate over longer distances (up to 15-20 km) (Paneque-Gálvez et al., 2014).

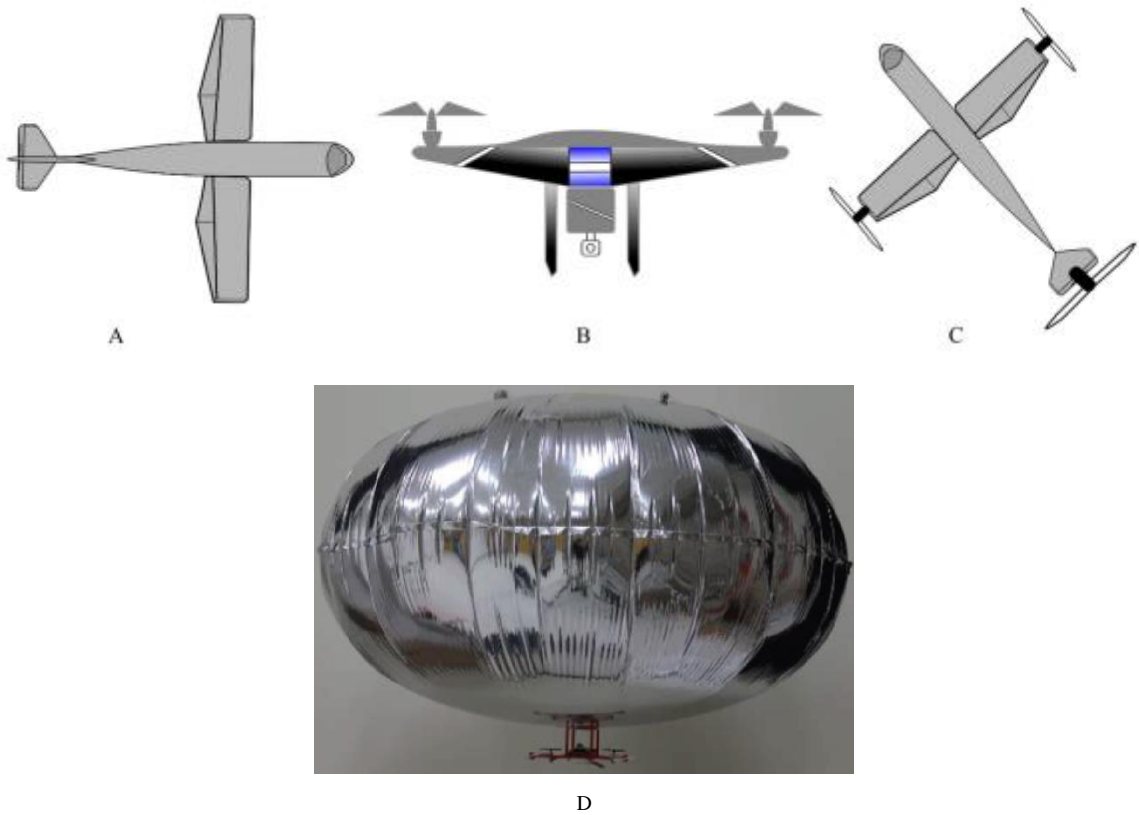

*Figure 1. Illustration of Basic UAVs (A) Fixed-Wing UAV (B) Rotary-Wing UAV (C) Combinational Concepts (source: Ahmad et al., 2021), (D) Blimps (source: Tao et al., 2018)*

On the other hand, rotary-wing UAVs is primarily categorized into the helicopter and multi-rotor types. The helicopter type of rotary-wing UAV has a unique feature with a large propeller atop the aircraft. It is widely used for spraying and aerial photography (see Figure 2) (Swain et al., 2010). They can hover, vertical takeoff, and land with nimble maneuverability while exhibiting low-speed flight for a shorter duration (Ahmad et al., 2021). In comparison, the multi-rotor models are called according to the number of rotors (Kim et al., 2019). Quadcopter, hexacopter, and octocopter are some multi-rotors UAVs that are widely known (Figure 3). These UAVs are lifted and propelled according to the number of rotors (Mogili & Deepak, 2018).

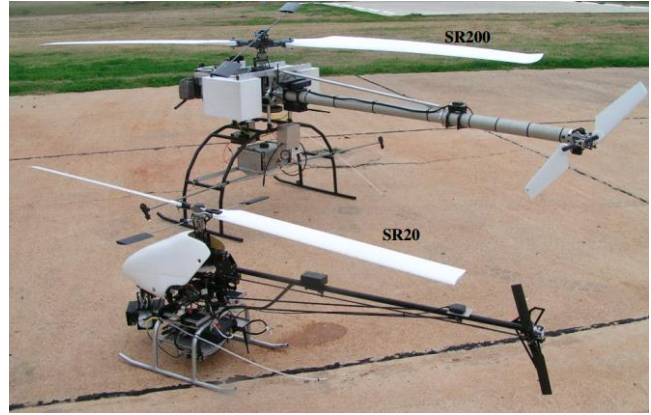

*Figure 2. Single Rotor/Helicopter UAV Type (source: Huang et al., 2009)*

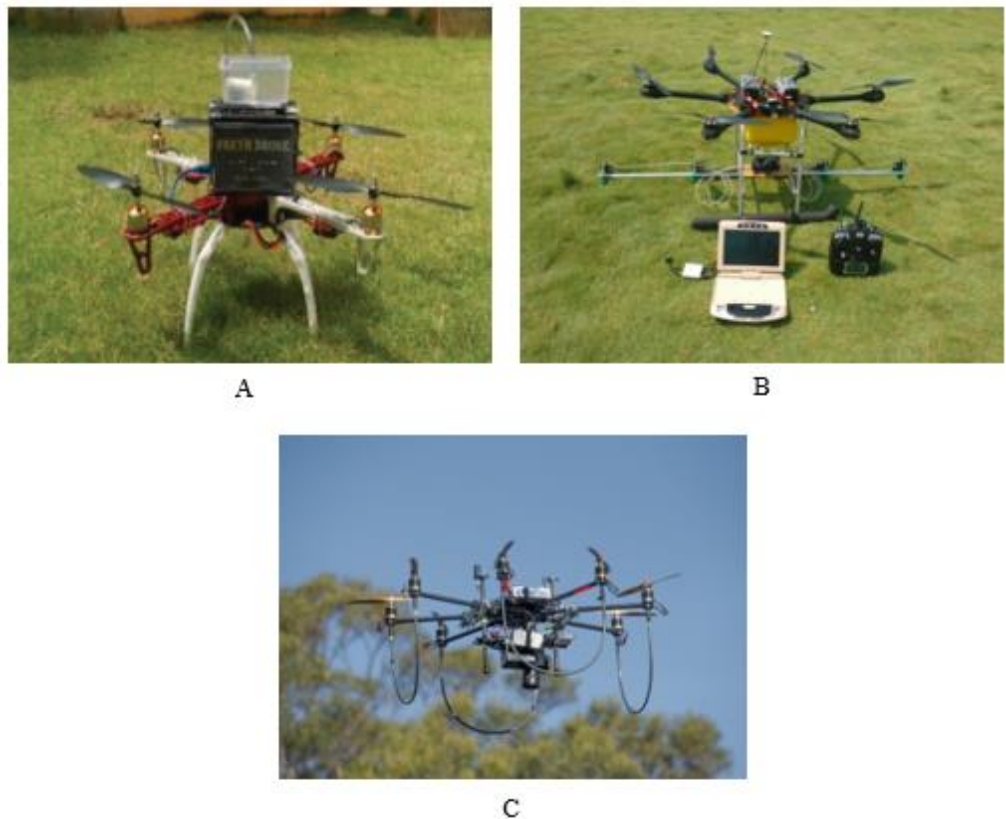

*Figure 3. Multirotor UAVs, (A) Quadcopter (source: Spoorthi et al., 2017) (B) Hexacopter (source: Yallappa et al., 2017) (C) Octocopter (source: Wallace et al., 2016)*

For example, the rotor movement of a quadcopter is responsible for generating the lift of a quadcopter. In a quadcopter, each of two rotors moves in an opposite way of which two rotors turn in the clockwise direction and the other two turn in the anticlockwise direction. The movement of the quadcopter around the axis consists of yaw (clockwise and anticlockwise), pitch (backward and forward), and roll angles (right and left). The quadcopter uses a control system to balance the thrust of each rotor in order to support the UAVs' lift and yaw, pitch, and roll angles (Mogili & Deepak, 2018). This control system turned out to be practical to produce a stable flight of the UAVs (Patel et al., 2013). Moreover, two quadcopter configuration types include the plus (+) and cross (X) models, as shown in Figure 4. The cross model is more popular between the two models due to its stability (Kedari et al., 2016).

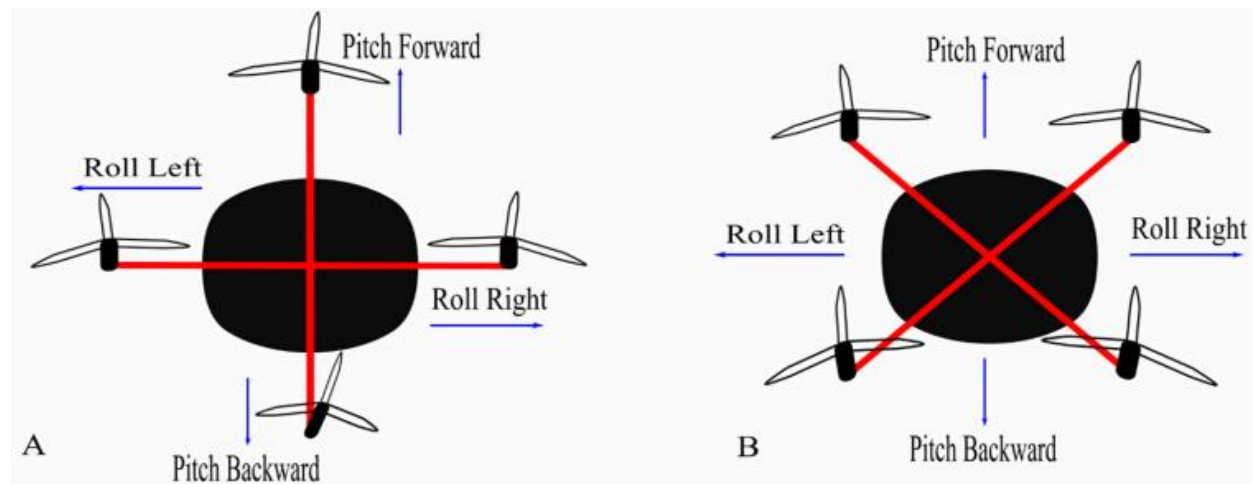

*Figure 4. Quadcopter Configuration Model (A) Plus configuration (B) Cross Configuration (source: Mogili & Deepak, 2018)*

Furthermore, these multirotor UAVs have extended their functionalities by equipping appropriate sensors such as vision, infrared, multispectral, and hyperspectral cameras. The expansion of UAV features brings great influences, especially in adding the capabilities of the UAVs. Those sensors are used to obtain data such as vegetation, reflectance indexes, and leaf areas in order to provide information about the current state of crops. With this information, farmers can make possible remedies or policies (weed control, fertilization, irrigation) according to the condition of the crops (Gonzalez-De-Santos, 2016).

Further, non-wing UAVs have been developed to cope with the long endurance of flying robots and are lighter than air (LTA) such as Blimp. A blimp is identically has a larger size than fixed-wing and rotary-wing and cushioned with a helium-filled envelope, making the robot safe to fly indoors, causing no threat to humans and the surroundings even with collisions (Tao et al., 2018). With the lifting force provided by air buoyancy, the blimp has flight endurance for more than 2 hours (Cho et al., 2017). Blimp is the one type of UAV lighter than the air UAVs (Krishna, 2021a; Thusoo, 2021; Tsouros et al., 2019). It has a balloon-like body created from tough fabric and filled with helium gas (Prisacariu et al., 2019). Blimp is notorious as the dirigible and was firstly designed in 1852 by Henri Giffard (Krishna, 2021a). This type of UAV has high endurance and can flow longer than other types of UAV, approximately 1-3 weeks travel (Krishna, 2021a). Due to its characteristics, the blimp is advantageous in numerous aspects of life, including the military and agriculture. It was purposeful as the cargo transit and the sentinels between the missile site and the military camp (Krishna, 2021a). It is also used to monitor the long-distance aspect, especially in urban traffic and buildings. In PA, it monitors crop production, identifies the plant's disease, erosion, and detects either flood or drought conditions (Krishna, 2021a). Unlike the other UAV, the blimp is also considered a safe technology since it remains in the air and did not collide even if it loses its power (Krishna, 2021a; Tsouros et al., 2019). Besides, the University of Leeds research reveals that blimp is chosen as the cheapest UAV to conduct a terrain survey (Krishna, 2021a). Thus, it could provide a detailed but further explanation of crops, surface minerals, vegetation, and water quality.

**4. Application of Blimp in Agriculture**

Compared to other UAVs, the blimp has a pivotal function, including load capacity, safety, quality, and environmental safes, which make it useful for everyday life. Researchers stated that blimp could carry up to 400 tons of load with 110-160 km speed travel (Krishna, 2021a). Besides its ability to fly longer, blimps could land on every land's surface. In the case of environmental footprint, blimp could reduce carbon dioxide emission (Krishna, 2021a).

Nowadays, there are several features of blimps with specific purposes. Those five types include tethered, untethered, remote-controlled blimp, Giga blimps, and hybrid blimps (Krishna, 2021a). Tethered blimp (aerostats) ables for free flight and a steady anchored flight using solid tethers (Mahmood & Ismail, 2020). A tethered blimp enables accurately obtaining a stereoscopic image that might cover 35 m2 to 20,000 m3 (Krishna, 2021a). Meanwhile, the untethered is commonly used in cargo transit, travel, and aerial surveillance. Thirdly, the remote-controlled blimp uses the robotic that use the program flight plan to fly. There are two different types of remote control blimps; small and large. The small blimp is commonly used for advertisement, and the giant blimp (Giga blimp) is significant for military purposes. Lastly, the hybrid blimp, a new modification of blimp that is prone to extreme conditions, can transport goods and civilian travel.

Unlike the other popular UAVs, multirotor and fixed-wing, the number of blimps applied in PA is insignificant (Krishna, 2021b; Tsouros et al., 2019). However, considering its strength characteristic and function, blimp could start considering the blimp as the priority to improve the

PA. (Mogili, Rao Deepak, 2021) stated that the integration of blimp with quadcopter aerial automated pesticide sprayer (AAPS) is pivotal for pesticide spraying in lower altitudes by following the GPS altitude. This technology is controlled with an android app to create an effective cost-saving (Mogili, Rao Deepak, 2021). Besides, other researchers use the blimp with a Charge-Couple Device to identify the Leaf Arena Index and biomass in soybean and paddy fields (Chilonga & Kiswisch, 2016). The results showed that the technology is stable and provides high-resolution images (Chilonga & Kiswisch, 2016). (Ponti et al., 2016) also stated that the blimp could be practically used to monitor the bean crop dataset using the combination of 1/2.3 inch of CCD sensor, 6.3 to 18.99 lens focal, and 10 Mega Pixel digital camera. The research found 29,556 examples of the positive dataset and 11404 negative datasets in Brazil (Ponti et al., 2016).

The blimp could be significantly used in monitoring agriculture (Mahmood & Ismail, 2020). For instance, the research conducted by (Bajoria et al., 2017) proposed a tethered aerostat system that could be used to mitigate the vertebrate mammal and bird hazard, which is positively contributed to 18%-43% of crop loss in India. The proposed design has been proven to carry about a 50 kg payload and 25 m/s ambient wind speed (Bajoria et al., 2017). Other research revealed that tethered aerostat combined with the electro-optical, acoustic, and laser-based sensors could scare the bird and other pests (Krishna, 2021a). To mitigate the occurrence of pests, the other researchers also create a Hawk Kite and Helikite aerostat hybrid that is purposeful to scare some of the bird's species, including the pigeons, seagulls, parrots, rooks, blackbird, etc. (Perigrine Ltd. 2018).

Besides, a tethered blimp (aerostat) could provide aerial images surrounding the natural disaster zone. This image helps identify the cropped field due to the flood, large soil erosion, drought, and crop loss due to the pest attack (Krishna, 2021a). It is also used to maintain the field quality because the aerostat could lofty 24 hours surveillance above. Thus, it could help the farmer control and watch the field without going directly to the farm. Besides the aforementioned reasons, the aerostat could reduce the enormous cost of capturing the crop's data. The farmer could use the aerostat to pertain the crop data and send its digital data in a computer program (Krishna, 2021a). Furthermore, the integration of aerostat with the sensor could help the farmer obtain continuous data of Nutrients crop status. Thus, it could help the farmer evaluate the number of nutrients placed for the crop (Krishna, 2021a).

**5. Agricultural Unmanned Aerial Vehicle**

In recent years, the application of more advanced technology in agriculture has gained more attention. Several technologies such as satellites, UAVs, Geographic Information System (GIS), Global Positioning System (GPS), and many other applications of technologies have been able to pave their way into the agricultural field. The process modernization and industrial revolution that brought many innovations in technology applications have opened the gate of precision agriculture (Ahmad et al., 2021). Precision agriculture is defined as the utilization of technology in the agricultural production system in order to determine, analyze, and manage the farming factors to increase crop productivity, ensure environmental sustainability, and improve business profitability (Unal & Topakci, 2013). This precision agriculture is seemingly possible to increase food production due to its effective functionalities under pressure conditions such as the ongoing reduction of arable land, the increase in global population, and the high cost of farming due to wastage in the use of water and chemicals (Abdullahi et al., 2015).

UAVs have gained popularity as a pivotal part of precision agriculture to ensure agricultural sustainability (Rani et al., 2019). The use of UAV, which plays a key role in reducing the data acquisition time and processing cost, is considered as the main reason for its popularity (Berni et al., 2009). The rapid development of UAVs that extend its functions to aerial photography and video and weather forecasting, with the support of spatial data collection to help stakeholders create policies and decisions, has attracted many parties to UAVs (Sylvester, 2018).

Moreover, the market of UAVs that is estimated to reach up to US$200 billion by the end of 2020 has successfully described the popularity of UAVs as well (Puri et al., 2017). The huge estimation of the total market of UAVs has shown that the market value of UAVs has doubled within three years. PwC's Drone Powered Solutions team quantified that the total market value of UAVs is about US$127.3 billion in 2017 (Silver et al., 2017). Although the estimation of the market value of UAVs in 2020 and the total market value of UAVs in 2017 are not exclusively focused on agriculture sectors, this number was sufficient to portray the UAV market development.

In addition, the affordable cost of UAVs is another factor that influences its popularity nowadays. This low-cost factor motivates many small companies to switch to using UAVs with its simple and easy-to-understand operating systems in serving some activities in the agriculture sector, including area measurement and crop monitoring (Hatfield & Prueger, 2010).

## 6. Applications of Unmanned Aerial Vehicles in Precision Agriculture

Currently, there are numerous applications of UAVs in precision agriculture. They are used in many areas of crops. This section introduces two applications of agricultural UAVs i.e., spraying and crop monitoring. The summary is shown in **Table** 1.

**6.1 Spraying**

Prior to the implementation of UAVs for spraying, the farmers used spraying bags to spray pesticides all over the farm (Spoorthi et al., 2017). Manual spraying is very dangerous for the workers because the measure of pesticides per hectare of agricultural land correlates to the risk of worker ailments. The heavy bag carried by the farmers could also make them get strained. Fortunately, the use of UAVs can reduce the usage of pesticides, maximize efficiency, and improve the well-being of the workers (Luck et al., 2010; Pyo, 2006).

Manual spraying is also considered ineffective for spraying the farmland because the pesticides may not spread evenly in every area. The excessive use of chemicals or pesticides in certain agricultural land is responsible for loss of soil fertility, soil degradation, and subsequent degradation of water-related ecosystems. In addition, the chemicals or pesticides absorbed by the crops and natural resources such as water and soil might cause pollution risk and severe health impacts for the environment. Therefore, UAVs are required to minimize such dangers by helping the spraying process specifically in the targeted area (Daponte et al., 2019).

In addition, to pave the way towards sustainable agriculture, the employment of UAVs in the agriculture sector also offers other benefits in terms of their operation. The implementation of UAVs can make the process relatively faster and cheaper than other methods (Rani et al., 2019). The efficient usage of UAVs was also widely reported in the literature. The use of 3WWDZ-10A, XAG is successfully effective in controlling *Spodoptera frugiperda*, an invasive sugarcane crop pest, by spraying pesticides (Song et al., 2020). In addition, the use of UAV (DJI Phantom 3) is found to be effective in spraying pesticides in the nominated areas using electronic traps (E-traps), which can count the insect and transmit the data to the server (Psirofonia et al., 2017). Studies have also found that UAVs might improve the accuracy of control over crops by equipping the UAVs with precision control algorithms (Faiçal et al., 2016). In summary, it is proved that the UAV application offers several advantages in reducing the workload of the farmers and providing efficient and low-cost service in the agriculture field.

Some issues have been identified regarding the use of UAVs in crop areas overlapping and outer edges, despite the advantages during UAVs implementation. These issues arise because some crop fields are not fully covered properly during spraying, leading to reduced crop quality in particular areas. To overcome this problem, the swarm of UAVs was introduced in a control loop algorithm during UAV operation (Yao et al., 2016). Swarm control is considered a practical technology since it could control multiple UAVs via one operator or program. In the swarm control method, the operator can select an efficient shape based on the application so that the swarm can be centralized, decentralized, and distributed according to the desired shape (Kim et al., 2019). In addition, the spraying pesticides process on the crop is then organized by considering the feedback from the Wireless Sensor Networks (WSNs) deployed in the field (Costa et al., 2012). The control loop is responsible for the communication of each UAV in adjusting the UAVs' route according to the changes in wind speed and the number of messages exchanged in between (Faiçal et al., 2014). During this communication process, a short delay might exist in the control loop since the UAVs need time to analyze the data from WSN to route further (Kale et al., 2015). An automatic navigation spraying system of UAV was developed to direct the UAV in a particular area (Xue et al., 2016).

Another way in using swarm control is through task allocation technology. This technology is currently used in mapping agricultural lands (Barrientos et al., 2011). In order to use swarm technology, a route is assigned to each UAV. A route is built by dividing each region or area among several UAVs. A map of the area is obtained by capturing a single picture through a camera sensor attached to a UAV (Ju & Son, 2018). This kind of technique requires K-mean algorithms in order to reduce complexity and prevent collision among UAVs. The most significant aspect of this swarm technique is the combination of algorithms that come in handy in maintaining the consistent distances between UAVs. These consistent distances allow linear and nonlinear control that resist strong external influences (Kim et al., 2019). The implementation of swarm techniques and task allocation in agriculture can be seen in Figure 5. This application most likely improves the accuracy of agricultural operations, reduces operator control efforts, reduces work time, and induces battery and payload shortages.

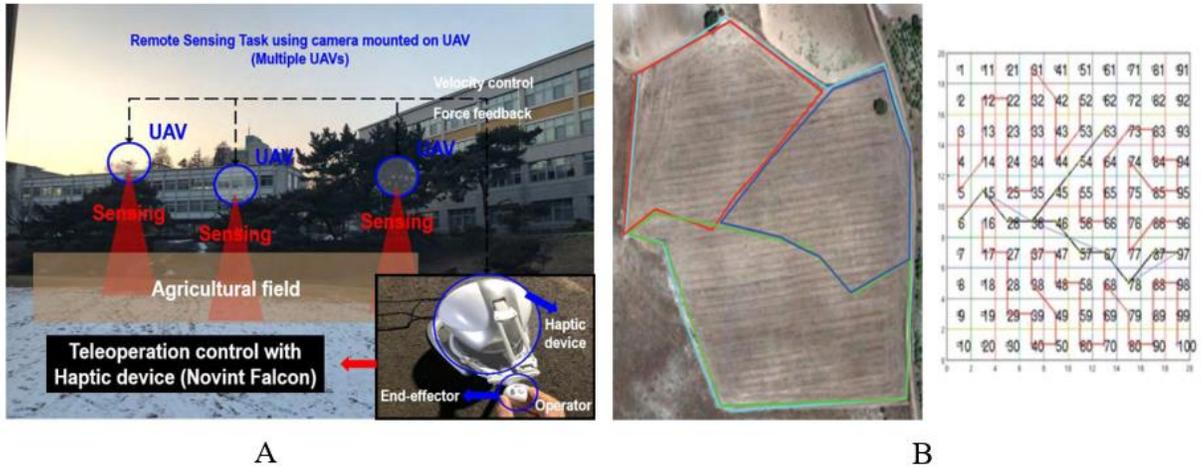

*Figure 5. (A) Swarm Control (source: Ju & Son, 2018) (B) Task allocation (source: Barrientos et al., 2017)*

In the spraying using UAVs, the sprinkling system is mounted at the lower region of the UAV, which has a nozzle under the pesticide tank in order to sprinkle the pesticide downstream in the field. An appropriately selected nozzle is a significant part of pesticide application since it is a significant factor in determining the amount of spray applied to an area, the coverage obtained on the target surface, the amount of potential drift, and the uniformity of application (Ru et al., 2014). Furthermore, the sprinkling system generally has two modules: the controller and the sprinkling system. The sprinkling system consists of the spraying content, either pesticides or fertilizers. Meanwhile, the controller is used to trigger the nozzle of the sprayer. The controller efficiency could be increased by using a PWM controller in pesticide applications (Zhu et al., 2010; Huang et al., 2009).

Another important component of the sprinkling system is a pressure pump used to put pressure into the pesticide in the tank to flow through the nozzle (Tang et al., 2018). This pressure pump works closely with the motor driver integrated circuit in completing their task in putting the pressure to sprinkle the pesticide (Mogili & Deepak, 2018). The full spraying system can be seen in Figure 6. The integration between UAV and the spraying system is expected to provide a potential platform for pest management and vector control, an accurate site-specific application for a large crop field. For this objective, a heavy lift of UAVs is required to cover many areas (Sarghini & De Vivo, 2017).

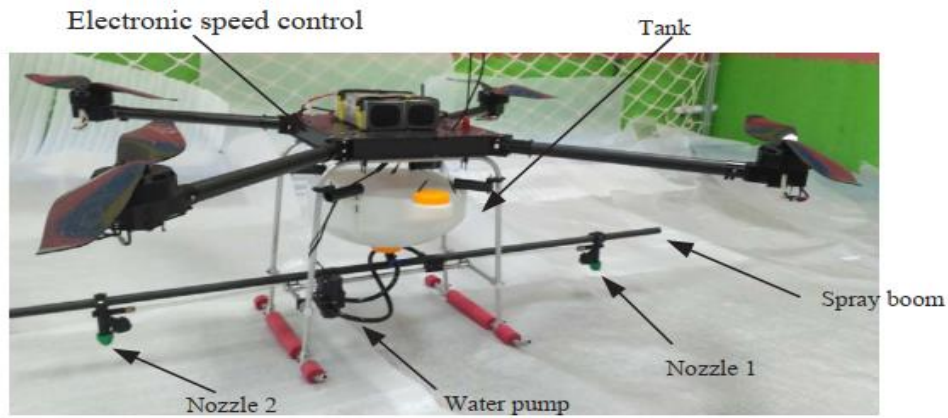

*Figure 6. Spraying System Structure Diagram (source: Tang et al., 2018)*

**6.2 Crop Monitoring**

A crop monitoring is defined as predicting the yield or crop quality by analyzing the available crop data (Kim et al., 2019). It is essential for optimizing crop production because it can assess crop health and indicate bacterial or fungal infections. Furthermore, the crop scanning produced by visible and near-infrared (NIR) light could reflect the different amounts of green light and NIR light that are extremely essential in producing multispectral images that can track changes in crops assess their health (Costa et al., 2012). The farmers can plan and apply remedies more precisely according to the identified issues with such information. It makes the fast response to bacterial or fungal infection, and infestation comes in handy and increases crop endurance into future issues.

The use of UAVs for crop monitoring is also highlighted due to their ability to monitor a large farm. By utilizing the UAVs, a large area of farmland can be fully monitored. It reduces the significant time and labor required for monitoring large farm areas manually (Kim et al., 2019). Aasen et al. (2015) reported that the UAVs application offers low crop monitoring costs. This is due to the use of lightweight sensors and the implementation of low-flying UAVs (see Figure 7).

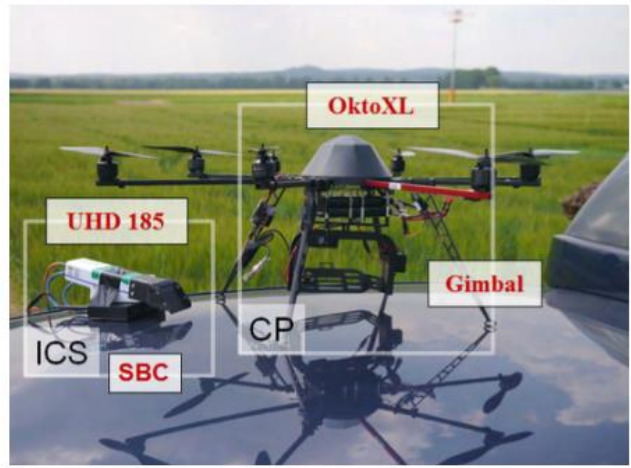

*Figure 7. UAV Platform (source: Aasen et al., 2015)*

A camera-equipped UAV can also observe the crop with different indices (Simelli & Tsagaris, 2015). Turner et al. (2011) used multispectral cameras mounted in UAVs to analyze the vegetation index of grapes obtained from vineyards. These vegetation index data are considered very important to emphasize the significant indicators to increase productivity and improve the shortcoming from farming activity. Furthermore, the application of UAVs in crop monitoring could also be seen through UAVs' capability to fly up to hectares of a field in one single flight. For this purpose, multispectral and thermal cameras are mounted at the UAVs' downside to recording the vegetation canopy's reflection (Bendig et al., 2012; Colomina & Molina, 2014). These cameras can take one capture per second and store it in the memory. The images are captured in the visible five bands with five different wavelengths (i) blues wavelength 440-510 nm (ii) green wavelength 520-590 nm, (iii) red wavelength 630-685 nm, (iv) red edge wavelength 690-730 nm, (v) near-infrared wavelength 760-850 nm. Then, those images were sent to the ground station through telemetry. The process of communication used the MAVLINK protocol. The data collected from the multispectral camera was analyzed by the geographic indicator Normalized Difference Vegetation Index (NDVI) (Reinecke & Prinsloo, 2017; Bhandari et al., 2012).

Moreover, the application of UAVs for crop monitoring has been implemented for conducting several tasks including monitoring crop growth, chlorophyll, and phenology measurement, and counting plants (Pino, 2019). These tasks are performed using SenseFly's e Bee Ag that has NIR and NDVI sensors. These sensors can replace traditional farm scouting by minimizing human error (Natu & Kulkarni, 2016). In addition, UAVs are involved in monitoring crops in hilly areas that are considered to be difficult for traditional scouting (Rani et al., 2019).

*Table 1. Applications of Agricultural UAVs*

| Task | UAV Model | Indices | Crop | Flight Altitude (m) | Sensors Type | Sensors Model | Task Period | Reference |
|---|---|---|---|---|---|---|---|---|
| Spraying | Fixed-wing UAV | Normalized Difference Vegetation Index (NDVI) | Maize Silage | 150 | - | Canon s110 | Throughout the year | (Castaldi et al., 2017) |
| | Helicopter | Spray Work Rate | Vineyard | 3-4 | Digital Camera | - | May | (Giles & Billing, 2015) |
| | | Route Precision, Spraying Uniformity | Wheat | 5, 7, 9 | Image Transmitter | - | Summer | (Xue et al., 2016) |
| | | Droplet size, Flow rate | Field | 6 | Proprietary Radio Receiver | - | Throughout the year | (Huang et al., 2009) |
| | | Leaf Area Index (LAI), Normalized Difference Vegetation Index (NDVI) | Maize Silage | 35 | Multi-Spectral Camera | Agrosenso | Throughout the year | (Castaldi et al., 2017) |
| | | - | Field | 20 | Wireless Sensor Networks | - | - | (Faiçal et al., 2017) |
| | Quadcopter | Time of Communication between a Sensor | Soy, Rice, Corn Gapes, Sugarcane | 5, 10, 20 | RF Module | XBee-PRO series 2 | Summer | (Faiçal et al., 2014) |
| | | Droplet coverage rate, Density, Droplet size | Cocktail, Grapefruit, Citrus | 3.5, 4, 4.5 | Digital Plant Canopy Imager | Camas CI-110 | Spring-Summer | (Pan et al., 2016) |
| | | Observed Deposition Rate, Field Work Rate | Field | Few meters | Multi-Spectral camera, Hyper-Spectral camera, Near-Infrared, Color-Infrared | - | Throughout the year | (Meivel et al., 2016) |
| | | Droplet Coverage Rate, Density, Droplet size | Citrus | 0.6, 1.2, 1.8 | Water-Sensitive Paper Cards (WSPs) | - | One day | (Tang et al., 2018) |
| | Hexacopter | Discharge and Pressure of Spray Liquid, Spray Uniformity, Spray Liquid Loss, Droplet Size and Density. | Paddy and groundnut | 1 | HD FPV camera | - | Throughout the year | (Yallappa et al., 2017) |

*Table 1. Cont.*

| Task | UAV Model | Indices | Crop | Flight Altitude (m) | Sensors Type | Sensors Model | Task Period | Reference |
|---|---|---|---|---|---|---|---|---|
| Crop Monitoring | Fixed-wing UAV | Normalized Difference Vegetation Index (NDVI) | Arable crops (corn, cotton, sunflower) | 120 | Multi-Spectral Camera | Parrot Sequoia Plus | June-October | (Bollas et al., 2021) |
| | | Normalized Difference Vegetation Index (NDVI) | Rice | 20 | Multi-Spectral Camera | Tetracam ADC camera | 95 days | (Swain et al., 2010) |
| | Quadcopter | NDVI, Ontario Soil and Crop Improvement Association | Soybean, Wheat, Barley, Oat, Canola | 120 | Digital Camera | Aeryon Photo3S | Spring-Autmn | (Zhang et al., 2014) |
| | | Visible-Band Difference Vegetation Index, Normalized Green-Blue Difference Wheat Index, Green-Red Ratio Index | Wheat | 100 | Digital Camera | SONY ILCE-6000 | September-July | (Du & Noguchi, 2017) |
| | | Leaf Area Index (LAI), Total Dry Weight (TDW), Plant Lenght (PL) | Three Rice Cultivars: Nipponbae (Japonica), IR64 (Indica), Basmati370 (Indica) | 30 | RGB Camera | Zenmuse X4s | Summer | (Peprah et al., 2021) |
| | | Vegetation Index (VI), Leaf Area Index (LAI) | Coffee | 30 | Digital RGB Camera | Sony EXMOR 1/2.3" | Throughout the year | (Barbosa et al., 2021) |
| | | Soil-Adjusted Vegetation Index (SAVI), Leaf Area Index (LAI), Normalized Difference Vegetation Index (NDVI) | Sunflower | 75 | Digital Camera | Tetracam ADC Lite | four days | (Vega et al., 2015) |
| | | - | Field | - | RGB Camera | - | - | (Doering et a., 2014) |
| | Hexacopter | Normalized Difference Vegetation Index (NDVI) | Vineyard | 150 | ADC-Lite Camera | Tetracam ADC-lite camera | One day | (Primicerio et al., 2012) |

*Table 1. Cont.*

| Task | UAV Model | Indices | Crop | Flight Altitude (m) | Sensors Type | Sensors Model | Task Period | Reference |
|---|---|---|---|---|---|---|---|---|
| Crop Monitoring | Hexacopter | Normalized Green-Red Difference Index (NGRDI) | Pea, Oat | 30 | RGB Camera | Panasonic Lumix DMC-GF1 | April-August | (Jannoura et al., 2015) |
| | | Blue Green Pigment Index 2 (BGI2), Reformed Difference Vegetation Index (RDVI) | Barley | 30 | Hyper-Spectral Camera | Firefly ultra-high definition 185 | Summer | (Aasen et al., 2015) |
| | Octocopter | NDVI, Soil Adjusted Vegetation Index (SAVI), Optimized SAVI (OSAVI) and Li | Barley | 50 | RGB-Sensor | Panasonic Lumix GXI | April-July | (Bending et al., 2015) |
| | | Structure-from-Motion (SfM), airborne laser scanning (ALS) | Eucalyptus Pulchella | 30 | RGB Camera | Canon55D | One day | (Wallace et al., 2016 |
| | | Normalized Difference Vegetation Index (NDVI), thermal temperature | Sugarbeet | 55 | Multiple Camera Array (MCA) Camera | Tetracam mini MCA | One day | (Bendig et al., 2012) |
| | | - | Sunflower | 122 | Multi-Spectral Camera | ADC Snap | - | (Noriega & Anderson, 2016) |

# 7. Limitations in Adopting UAVs Technologies in Agriculture Sector

## 7.1 Technical Decisions

Various types of UAVs have been produced in the commercial market by many manufacturers and companies, starting from hobby-type products up to industrial model aircraft. Since there is no specific standard about the UAV development for agricultural purposes, it is hard to find a UAV built specifically for the agricultural context (Huang et al., 2013). Moreover, suppose the available commercial software packages, which support the photogrammetric data processing, are not standardized for agricultural purposes. In that case, the desired UAV images may not be appropriately captured by the sensor. Therefore, it can prevent the users from taking the right actions if unexpected situations such as a collision with another flying object occur (Abdullahi, 2015).

Another major problem associated with technical decisions is the battery usage and flight time limitations. The lithium-ion batteries currently used in UAVs have an advantage over conventional batteries, especially in their larger capacity. However, the larger capacity affects the weight of the batteries that become heavier in return (Saha et al., 2011). Unfortunately, this issue is challenging to be solved within this day. Another problem related to battery usages is battery management. Even though it is known that the batteries of UAVs need to have constant maintenance, most UAV operators often forget and do not carefully pay attention to this issue. As a result, it caused periodic replacement that required additional cost (Lee et al., 2012). Lastly, the possible time for UAVs to fly, which is around 20-30 minutes with a fresh battery, can still provide enough time for complete crop monitoring (Baha et al., 2012). Researchers try to develop optimized hybrid batteries as solutions in dealing with this issue.

**7.2 Cost**

The lack of awareness of the UAVs' cost, is one of the reasons for the slower adoption of this technology in the agriculture sector. For a starter system, agricultural UAVs can range from US$1,000 that might go up to US$10,000 or US$20,000, depending upon the cameras and the features (Stehr, 2015). This cost is not quite affordable and surely will be an impending stop to adopt UAVs technology for smallholder farmers (Ahmad et al., 2021). The interested farmers who could not afford the cost of UAVs may need to contract as a group to get UAV services to reduce the individual expenses.

Another possible solution to minimize the cost of UAVs by purchasing inexpensive airframes and low-cost cameras. However, this solution could build up a short endurance of the UAV platform. Moreover, the low-cost UAVs are usually equipped with lightweight engines that might limit the reachable altitude of the UAVs. The low cost of cameras also limited the sensor payload both in dimension and weight, and reduced image quality (Abdullahi, 2015). In addition, the separate purchases of UAV components require highly skilled engineers or technicians to integrate and assembly, which may increase the total expenses (Huang et al., 2013).

Apart from the cost of vehicles equipped with cameras and software for aerial imagery processing, the farmers need to consider the expenditures for the operator's license. The presence of this operator implies extra time and cost that need to be spent since not everyone is allowed to operate the UAVs. Nevertheless, all these costs will constantly decrease over the years (Bollas et al., 2021).

### 7.3 Payload

Payload weight and size are critical factors for UAVs because they need to be carefully configured based on the specific application of the UAV. When the UAV is ready to use, it needs to be configured by paying attention to payload design, mechanical and electrical accommodation even though there is no specific engineering guideline to be followed (Huang et al., 2013).

### 7.4 Operation

In the UAV operation, most UAV types do not have the capability of automated take-off and landing (Huang et al., 2013). Furthermore, the frequency of flying UAVs should be carefully selected because there are insufficient regulations about flying UAVs. Even certain regions restrict the usage of UAVs as a security precaution (Eisenbeiss, 2009). Another challenge is the UAVs' inability to take readings during extreme weather conditions like rain or storm (Abdullahi et al., 2015). Therefore, highly skilled operators for remote control are required. However, the demand for skilled users to operate the UAVs is a problem for small and medium producers to adopt UAV technology. Training issues and lack of demonstrated financial returns in the short and medium term are considered the reason for this issue (Abdullahi et al., 2015). Thus, autonomous flight according to georeferenced coordinates has then become a highly desirable component for practical use of UAVs in agriculture (Huang et al., 2013).

The swarm-control techniques can be applied to efficiently control multiple UAVs in performing a wide range of tasks. Although swarm-control can provide practical techniques to lower the battery cost and operate more efficiently with shorter flight times, there is a need for user interface improvement so that people who are older or unfamiliar with UAVs can easily control the UAVs. The user interface improvement is made by considering multimodal feedback, including visual, auditory, and haptic feedback. Therefore, an improvement that mainly focused on human-centered user interface and feedback are two ways that seem to be effective to deal with multiple UAVs (Hong et al., 2017).

**8. Challenges in Implementing UAVs in Indonesia**

Kavianand et al. (2016) have reported that agricultural development in Indonesia is critical since it has primarily contributed to Indonesia's GDP. Roughly about 14.4 percent of Indonesia's total GDP comes from the agriculture sector and has reduced the unemployment rate by absorbing 38.6 percent of the workforce (David & Ardiansyah, 2017). Despite its considerable contribution to Indonesia's GDP, the contribution of agriculture to Indonesia's GDP has been remarkably decreasing for the last five decades due to low productivity. Some natural phenomena, such as extreme weather changes, have also influenced Indonesia's agriculture (Syuaib, 2016).

Many researchers have suggested implementing precision agriculture via UAVs to improve the productivity of the work in agriculture. The application of UAVs offered many benefits that could grow the economic profit and provide a proper solution in solving current issues in agriculture. However, as much as Indonesia depends upon agriculture, the application of UAVs in the agriculture field is relatively far from adopting the latest technology into farms. Even though some developed countries have started to use UAVs in their precision agriculture and proved that this technology is essential in reducing farmers' workload, Indonesia seems to fall behind and keep using manual operating for farm activity.

One of the viable reasons for preventing the adoption of UAVs is the education level of farmers. The majority of farmers in Indonesia do not complete their high school education in which 38 percent of local farmers have graduated from primary school (Haq et al., 2016). Furthermore, only 6 percent of the local farmers can complete high school or university. These numbers significantly describe the current state of Indonesia's farmers. The lack of education could cause a low understanding of the technology application. Moreover, this low level of understanding can lead to the anxiety of relearning integrating agriculture and technology (Suryanegara et al., 2019).

Despite the reasons mentioned above, researchers have tested the application of UAV in several agricultural sectors. For instance, the UAV has been practically implemented or tested in Indonesia's agriculture, including the sugar cane plantation in PTPN (Perkebunan Nusantara Maospati East Java) and a paddy field near Menara Cigarette Factory, and the Teak Wood Forest in Madiun (Rokhmana, 2015). Besides, the application of UAV has been significantly tested in one of the paddy fields in Parankasalak, Sukabumi, West Java, to monitor the crop by mapping the paddy field through differentiating them based on their spectral characteristic (Rokhmatuloh et al., 2019). However, the research found that the implementation of UAV poses several limitations and challenges that become a prohibitive factor for broader use in Indonesia's agriculture.

The application of UAV in PA requires a high investment cost to purchase the technology and the maintenance cost (Tsouros et al., 2019). Furthermore, due to the limited space of agricultural land and the unstable market price for the crop yield, the implementation of UAV might pose another operational cost to the farmers (Tsouros et al., 2019; Vera et al., 2021). Even though the market has commercially offered an amateur and cheaper UAV, the product has several limitations related to stability, accuracy, and quality (Norasma et al., 2019). The cheaper UAV has a low ability to reach a certain altitude due to its low power engine (Norasma et al., 2019). This notion is reinforced by (Rokhmana, 2015) who notes that amateur UAVs generally have an error in their camera lens. This case happens because both the stability and accuracy of the non-metric lens are low. Besides, the UAV is relatively light, possesses only 3 kg weight, which causes them to be easily disturbed by the wind and air turbulence when the weather is windy (> 40km/h) and rainy days (Norasma et al., 2019; Rokhmana, 2015; Tsouros et al., 2019). This case requires huge attention because Indonesia is located along the equatorial belt region to have periodic heavy rain.

Furthermore, the UAV requires data-intensive procedures and skilled personal for exploiting the acquainted imagery data. (Tsouros et al., 2019). Hence, the farmers need to hire the expertise of UAV technology or do intensive training that may be costly. This case requires intensive consideration because the average farmer in Indonesia is not in productive ages with low educational background (Haq et al., 2016). It reported that 88% of the average farmer in Bantarkawung is on the 15-60 years and the remaining farmer is on non-productive ages (Haq et al., 2016). Moreover, it discovered that only 6% of the majority graduated from high school and university; the remaining only attended primary school, and 38% were in junior high school (Haq et al., 2016). As a result, most farmers have a common understanding of technology, IoT and little comprehension of imagery data. Another viable reason for preventing people from using the UAV technology is its limited flight time. (Tsouros et al., 2019) revealed that most commercial UAVs only have 20 min to 1 hour flight time. Moreover, it only covers a small restricted area for each flight. Thus, the total cost expenses to purchase the UAV technology for PA might not be advantageous.

## 9. Conclusion

The application of UAVs in current days has opened unlimited potential, especially in the agriculture sector. Two main UAVs applications in agriculture sectors, such as spraying, and crop monitoring have been discussed. The urgency of UAVs and the implementation of UAVs were necessary to be implemented in order to establish precision agriculture. Numerous issues and problems that might occur in the future have also been highlighted to build awareness about the issues by providing various data and sources. The application of UAVs in spraying and crop monitoring are the main parts of this paper since we were thoroughly investigating the application of UAVs that includes the benefits obtained, various application forms of UAVs from several types of research, and the flow of operating the UAVs. Moreover, the limitation found in the application of UAVs was also identified to reveal the gap of UAVs implementation in the agriculture field. Lastly, the challenges in implementing UAVs are also being discussed, especially in Indonesia.